\documentclass[10pt, conference, compsocconf]{IEEEtran}

\usepackage{cite}
\ifCLASSINFOpdf
\usepackage[pdftex]{graphicx}
\else
\usepackage[dvipdf]{graphicx}
\fi
\usepackage{amsmath}
\usepackage{color}
\usepackage{amssymb}
\usepackage{subcaption}
\newcommand{\comment}[1]{}
\newcommand{\etal}{\textit{et al}.}

\makeatletter
\newcommand{\printfnsymbol}[1]{%
  \textsuperscript{\@fnsymbol{#1}}%
}
\makeatother

\begin{document}
\title{Serif or Sans: Visual Font Analytics on Book Covers and Online Advertisements\\[-5mm]}
\author{\IEEEauthorblockN{Yuto Shinahara\IEEEauthorrefmark{1},
Takuro Karamatsu\IEEEauthorrefmark{1},
Daisuke Harada\IEEEauthorrefmark{1}, 
Kota Yamaguchi\IEEEauthorrefmark{2} and
Seiichi Uchida\IEEEauthorrefmark{1}}
\IEEEauthorblockA{\IEEEauthorrefmark{1}Kyushu University, Japan}
\IEEEauthorblockA{\IEEEauthorrefmark{2}CyberAgent, Inc., Japan}
}


\maketitle
\begin{abstract}
In this paper, we conduct a large-scale study of font statistics in book covers and online advertisements. Through the statistical study, we try to understand how graphic designers relate fonts and content genres and identify the relationship between font styles, colors, and genres. We propose an automatic approach to extract font information from graphic designs by applying a sequence of character detection, style classification, and clustering techniques to the graphic designs. The extracted font information is accumulated together with genre information, such as `romance' or `business', for further trend analysis.
Through our unique empirical study, we show that the collected font statistics reveal interesting trends in terms of how typographic design represents the impression and the atmosphere of the content genres.
\end{abstract}


\IEEEpeerreviewmaketitle

\section{Introduction}
\begin{figure}[t]
  \centering
  \includegraphics[width=\columnwidth]{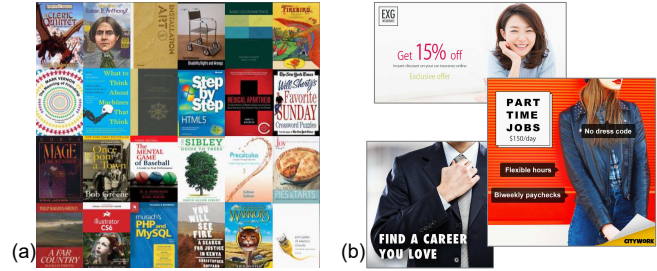}
  \caption{Examples of a) book covers and b) online advertisements. We propose to automatically extract, categorize, and analyze font information from these graphic designs.
  }
  \label{fig:bookcovers}
\end{figure}

Various visual designs, such as poster, book cover, advertisement, or product packaging, contain textual information printed in various fonts. Fig.~\ref{fig:bookcovers} shows examples of book covers and online advertisement graphics. \textit{Why is a certain text printed with a specific font?} For example, engineering documents might have a bold sans-serif title. This question can be rephrased as follows: {\it how typographers to choose a font for the book?} We can imagine that there is no strict and universal rule to visual designs, and the creation process can involve various social and psychological factors of individual designers. However, still there might exist general principles or tacit rules by designers that we could observe as trends in a large volume of visual design examples.

The purpose of this paper is to study the statistical trends of font usage in visual designs. Using two different large design datasets, i.e., book covers and online advertisements, we aim at undermining hidden trends in font use in graphic designs via computational approach. For analyzing font usages in book titles, we automatically extract a title part from book covers using scene text detector and identify its font style (e.g., Sans-serif) and font color. Our empirical study from 207,572 book covers reveals unique trends in font styles and colors for different book genres, such as the strong preference of decorative fonts in recreational books. Our cluster analysis suggests there are clusters of similar book genres in terms of font use, which often reflects the proximity of the topic in terms of content. Our second study using 30,000 advertising banners also suggests there exist unique trends of font usage at different business category. Our automated analysis successfully discovers hidden rules or common sense that typographers intentionally or unintentionally use.


To the best of authors' knowledge, our study is one of the first attempts in quantitative and automatic font analysis. Our approach builds upon recent methods of scene text detection and recognition techniques~\cite{zhou2017east,shi2017end}, and well demonstrates how a computational approach enables a systematic scientific study at a scale that has been impossible in the past. There have been psychological insights that graphic designers or typographers choose fonts based on impression and atmosphere of the content~\cite{brumberger2003rhetoric,henderson2004impression,doyle2006dressed,mackiewicz2007audience}, and our result provides quantitative evidence for existing hypotheses or verbalizes tacit rules in design. From an engineering perspective, our statistical analysis enables the development of interesting applications such as automatic font recommendation, font generation, or typography evaluation.


Our contributions are summarized in the following:
\begin{itemize}
    \item We propose a novel attempt in visual font analysis using recent text recognition techniques; and
    \item We conduct a large-scale empirical study using real-world datasets of 207,572 book covers and 30,000 online advertising and show there indeed exist specific patterns in font usage in visual design.
\end{itemize}


\begin{figure*}[!t]
  \centering
  \includegraphics[width=0.9\textwidth]{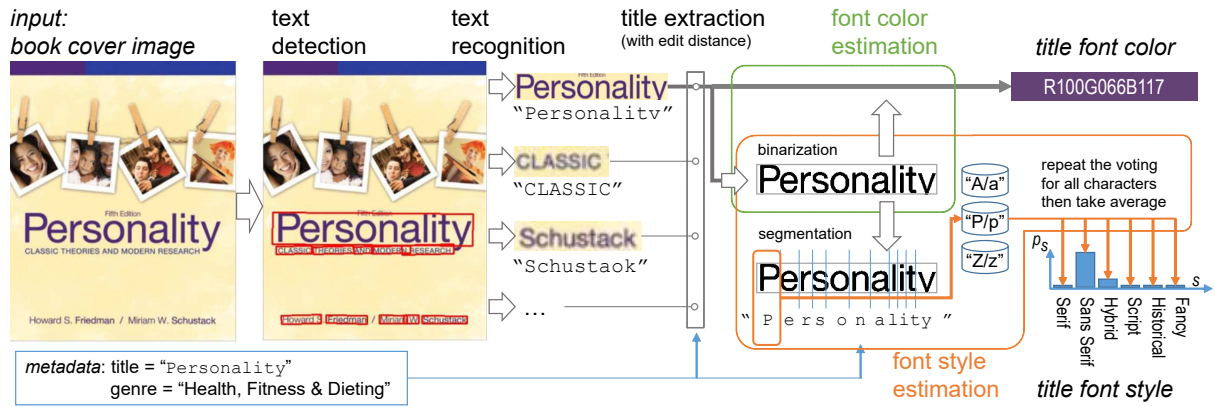}
  \vspace{2mm}
  \caption{Our pipeline for font style and color extraction from book titles.}
  \label{fig:flowchart}
    \vspace{-3mm}
\end{figure*}

\section{Related work}
\subsection{Design and Psychology Literature}
In design research, font usage has been dealt with mostly as subjective knowledge of typographers. While there are many typography guides and hints for appropriate font usage, they are rather intuitive suggestions and merely supported by any quantitative fact derived from data analysis. On the other hand, visual design analysis, such as advertisements~\cite{hussain2017automatic, ye2018advise} or infographics~\cite{bylinskii2017understanding}, is now a trend in the computer vision community. However, font usage in those visual designs has not been analyzed even in the trend.

In psychology literature, the perceptual impression of fonts has been a research topic from the 1920's~\cite{poffenberger1923study,davis1933determinants}. Following those pioneering
studies, many modern psychological studies have revealed the relationship between font style and impression~\cite{brumberger2003rhetoric,henderson2004impression,doyle2006dressed,mackiewicz2007audience},
and still the topic is actively studied~\cite{grohmann2013using,o2014exploratory,velasco2015taste}. Those studies reveal the impression of fonts through subjective human experiments with  questionnaire answering.
Among them, O'Donovan \etal~\cite{o2014exploratory} create a notable font dataset where 37 attributes, such as wide, complex, and friendly, are attached to the individual fonts.

In contrast to design and psychological literature, we take a computational approach to analyzing fonts from large collections of graphic designs. Our results are totally objective and reproducible. In addition, the results are reliable enough because they are derived from 207,572 book covers and 30,000 advertisement images.

\subsection{Computer Science Literature}
Font generation and font recognition have recently become a hot research topic in computer science.  
For font generation, a subspace representation~\cite{campbell2014learning}, a pairwise interpolation~\cite{uchida2015exploring}, and a patch-based optimization~\cite{yang2017awesome} are proposed. Recently, many generative adversarial networks (GAN)-based font generation methods (e.g., \cite{azadi2018multi, guo2018creating, jiang2017dcfont}) are proposed and applied even to huge-category tasks, such as Chinese font generation. For font recognition, CNN-based methods~\cite{wang2015deepfont, wang2018font} are proposed and showing accurate performance. For example, DeepFont~\cite{wang2015deepfont} achieves 80\% as top-5 accuracy on a 617 font-class task.

Recently, there are a couple of advanced attempts for generating or choosing an appropriate font to the input image. Zhao \etal~\cite{zhao2018modeling} propose a method that generates a font while treating a given Web-page image as a visual context. It also estimates appropriate color and size for showing the font. Choi \etal~\cite{choi2018fontmatcher} proposed a method that chooses a font while considering the atmosphere of the input scenery image.

In addition to those font studies, scene text detection and recognition methods are intensively studied so far. Recent methods based on deep neural networks can detect and recognize scene texts even if they are printed with heavily-decorated fonts. Those robust scene text detectors and recognizers are mandatory to extract ``fonts in the wild'' before their analysis. Among the recent methods, EAST~\cite{zhou2017east} and CRNN~\cite{shi2017end} becomes one of the standard methods thanks to their ground-breaking performance on scene text detection and recognition, respectively. In this paper, we also use EAST and CRNN for extracting fonts used for book titles from book cover images, because they showed sufficient performance for our task.

We emphasize that quantitative analysis of font usage is a novel direction of research in the document analysis and recognition community. There is no established state-of-the-art. In fact, it was almost impossible so far to collect many font usages without recent scene text detection and recognition methods introduced above. 

\begin{figure*}[!t]
  \centering
  \includegraphics[width=0.95\textwidth]{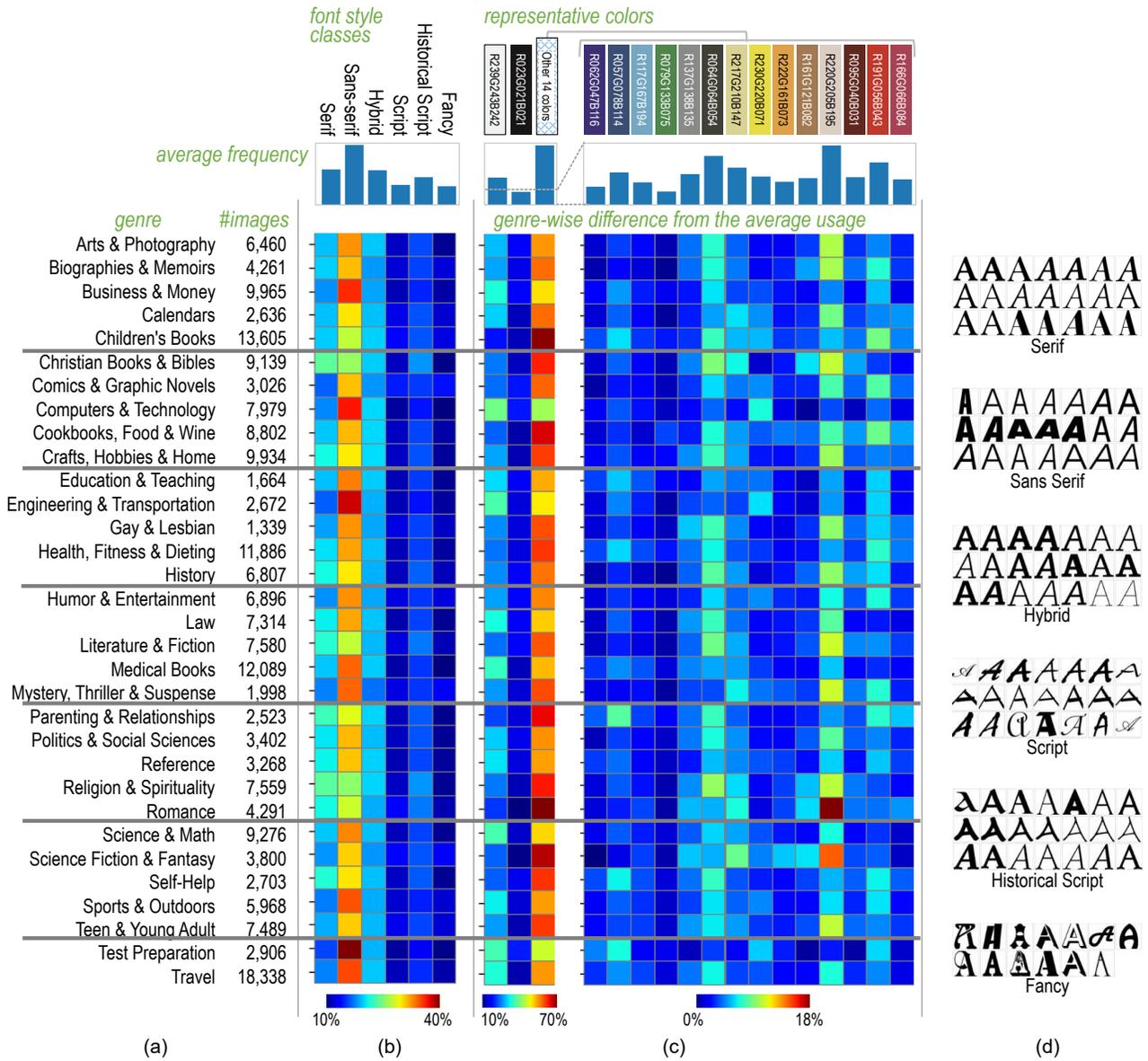}
  \caption{Summary and visualization of font-genre relationships in book covers: (a) 32 book genres and their associated number of images, (b) style usage, (c) color usage, and (d) font style examples.}
  \label{fig:hm-book}
\end{figure*}
\section{Font Analysis in Book Titles}
\subsection{Book Cover Dataset}
We collected 207,572 book cover images from {\tt Amazon.} {\tt com}. The image dataset is published at this URL~\footnote{\tt https://github.com/uchidalab/book-dataset}. The dataset consists of a collection of cover images and associated metadata such as a title or a genre. Fig.~\ref{fig:hm-book}~(a) lists 32 book genres defined by {\tt Amazon.com} along with the number of the cover images at each genre. Each book belongs to a single genre and the number of books varies according to the genre.

\subsection{Font Style and Color Extraction}\label{sec:fontcolor-book}
Fig.~\ref{fig:flowchart} shows our pipeline for identifying the title font from a book cover image. The steps are divided into three parts; title text extraction, title font color estimation, and title font style estimation.

\subsubsection{Title Text Extraction}
As the first step for title text extraction, we apply EAST~\cite{zhou2017east} to detect all words as bounding boxes from the book cover image. EAST is robust to non-horizontal layouts, font decoration, size variations, and color variations, and thus can accurately detect most words. Next, we use CRNN~\cite{shi2017end} to recognize each detected word. CRNN can be a lexicon-free word recognizer and thus has the potential to recognize even proper nouns and rare words. Finally, the bounding box of the title text is extracted from the recognized words using metadata of the book title. Edit distance is used here to identify correct words because there might be some character recognition errors from CRNN. We extract only title fonts and discards other irrelevant texts such as author names.
\subsubsection{Font Style Estimation}
We estimate font styles at individual characters in the extracted title text. Instead of estimating the exact font name (such as ``Helvetica''), we estimate the style category of the font. This is because an accurate font recognizer, such as DeepFont~\cite{wang2015deepfont}, is not publicly available and many rare fonts are used in the book titles.
In our setup, we categorize fonts into one of six font styles: 1) Serif, 2) Sans-serif, 3) Hybrid (mixture of serif and sans-serif), 4) Script, 5) Historical Script, and 6) Fancy, according to \cite{book}, where 1,132 Adobe fonts are classified into these styles.
We show font examples in Fig.~\ref{fig:hm-book}~(d).

%
A detailed procedure for font style estimation is explained in the following. Let us assume we are given that the title is ``Personality'' by metadata. First, the bounding box of the word image ``Personality'' is partitioned into individual character components by the Otsu's binarization. The resulting 11 character components are labeled as `P', `e', `r', `s', and so on. Then the image patch of the first character `P' is compared with all 1,132 font images of `P' and its $K$ nearest-neighbor font images are selected for each style $s$. The probability that the character`P' is printed in the style $s$ is given by $p_s = \sum_{k=1}^K (1/d_{s,k}) / (\sum_{s'=1}^6 (1/d_{s',k}))$, where $d_{s,k}$
is the pseudo-Hamming distance~\cite{uchida2015exploring} to the $k$-th nearest-neighbor font image of class $s$. The probability vector $p=(p_1, \ldots, p_s, \ldots, p_6)^T$ is averaged over all 11 character components of the title and then treated as the font style usage of the book. In the experiment, we set $K=3$ because there is no significant difference with a larger $K$.


\subsubsection{Font Color Estimation}
We estimate font color by separating the character strokes from their background within the bounding box. We separate stroke regions by Otsu's binarization. Next, we average L$^*$a$^*$b$^*$ colors of all foreground pixels and treat this as the font color. Although averaging is not appropriate when the title text has multiple colors, we find such a case is rare and almost negligible in our dataset.
The averaged color is finally quantized into one of the representative colors in the dataset. We use 16 representative colors (Fig.~\ref{fig:hm-book}~(c)) determined by $k$-means on all 207,572 L$^*$a$^*$b$^*$ color vectors.

\subsection{Results}
\subsubsection{Font Styles}
Fig.~\ref{fig:hm-book} (b) shows the frequency of font styles of all 32 genres as a bar chart and that of each genre as a heat map. The heat map of a genre is the average of the probability vectors $p$ over all book titles of the genre. Consequently, the sum of the heat map values of each genre is always normalized to 100\%.
We identify the following trends:
\begin{itemize}
    \item Sans-serif fonts are the majority of all book genres. Especially, those fonts are very frequently used at genres for practical books (how-to books), such as ``Business \& Money'', ``Computers \& Technology'', ``Engineering \& Transportation'', and ``Test Preparation.'' In other words, these genres use Serif fonts much less frequently than other genres.
    \item Serif and historical script are used in genres about humanity, such as   
    ``Literature \& Fiction'', ``Religion \& Spirituality'', and ``Christian Books \& Bibles.''
    \item Fancy fonts (i.e., decorative fonts) are not dominant but found at genres of recreational books, such as ``Comics \& Graphic Novels'', ``Mystery, Thriller \& Suspense'', and ``Science Fiction \& Fantasy.'' 
    \item Script and fancy appear as frequent as most other genres, although script and historical script have different trends. This suggests that script fonts are treated as decorated fonts rather than classical fonts.
\end{itemize}

\subsubsection{Font Colors}
Fig.~\ref{fig:hm-book}~(c) shows the frequency of font colors of all 32 genres as a bar chart and that of each genre as a heat map and reveals the following trends:
\begin{itemize}
    \item Achromatic colors (white, black, gray) and pale colors are the majority in overall.
    \item Vivid colors become dominant in some cases, such as red in ``Children's Books'' and yellow in engineering-related genres (``Computers \& Technology'' and ``Engineering \& Transportation'').
    \item Dark chromatic colors (blue, green) are not common, except for the usage of dark blue in genres relating to study  (``Parenting \& Relationship'', ``Education \& Teaching'', ``Test Preparation'' and ``Children's Books'') and health (``Health, Fitness \& Dieting'' and ``Self-Help''). 
    \item White is often used in genres for practical books.
    \item Pale pink is often used in genres for humanity and recreation books, such as ``Romance''.
\end{itemize}

\subsection{Discussion}
\begin{figure}[!t]
  \centering
  \includegraphics[width=\columnwidth]{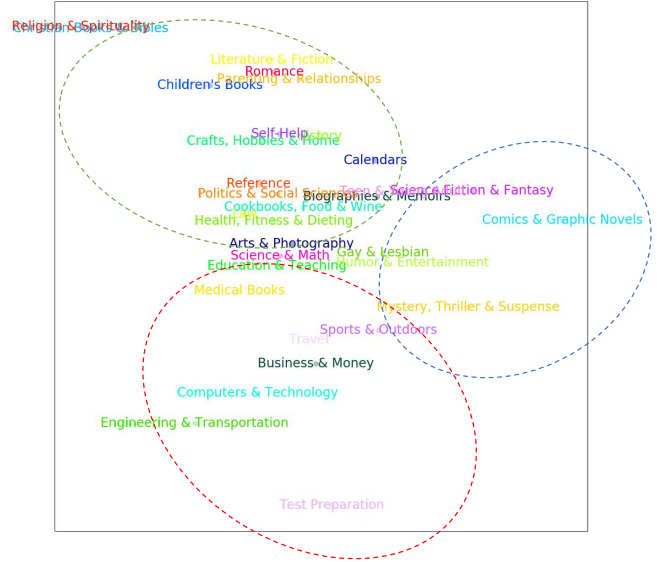}
  \caption{Proximity of book genres by font styles. Genres about practice, recreation, and humanity form clusters around red, blue, and green circles respectively.}
  \label{MDS}
\end{figure}
\subsubsection{Do Similar Genres Share Similar Font Styles?}
From the empirical observation, one may expect that similar genres have similar font style usages. This expectation is supported positively by Fig.~\ref{MDS}. We show in Fig.~\ref{MDS} a two-dimensional visualization of the genre proximity, which we obtain by applying multi-dimensional scaling (MDS) to the six-dimensional font style frequency vectors of all 32 genres. As suggested by circles, similar genres get close to each other; for example, genres for practical books form a cluster of the red circle.
These results prove that there are clear trends in font style usages for each genre, and typographers purposefully choose an appropriate font for a book title under these trends. We emphasize that those findings arise from our automated analysis of 207,572 book covers.
Note that we also had a similar genre distribution with font color usage, although it is omitted due to space constraints. 

\subsubsection{What Does It Mean to Be an ``Average'' Genre?}
According to the property of MDS-based visualization, we can say that genres around the peripheral part of the two-dimensional genre distribution of Fig.~\ref{MDS} have very unique usage of font styles. Examples include: ``Comics and Graphic Novels'', ``Test Preparation'',  ``Religion \& Spirituality'' and ``Christian Books \& Bibles.''  In contrast, genres around the center of the distribution
often imply that those genres have a very large variation and thus does not have any focus on a specific font style. For example, fine arts have various styles and subjects. Therefore, book titles in the genre ``Arts \& Photography'' are printed in various fonts and therefore the genre goes to the center of the distribution.

\section{Font Analysis in Online Advertisements}
\subsection{Online Advertisement Dataset}
As another example to analyze genre-wise font usage, we collected 28,098 online advertisement images from our industrial partner in Japan.
Fig.~\ref{fig:hm-adv}~(a) shows their seven business genres we pick from IAB categories~\cite{iab2017} and the associated sample size. There can be multiple graphic designs for the same product. The column ``\#products'' shows the number of products for each genre. On average, we observe $28,098/200\sim 140$ designs per product. Fig.~\ref{fig:hm-adv} (a) also shows the number of companies for each genre.

\subsection{Font Style and Color Extraction}
The ads have two different properties from book covers. First, we can get the precise font name (like ``Century Gothic'') and their color by inspecting a vector file format (Adobe Photoshop format). This property makes the analysis very simple; we need to perform neither text detection nor font style estimation. Second, there is no ``title'' in ads. We instead need to analyze all the texts in the banner graphics. In ads, texts appear in various sizes. We apply weight for each text proportional to its size in the analysis.

\begin{figure}[!t]
  \centering
  \includegraphics[width=\columnwidth]{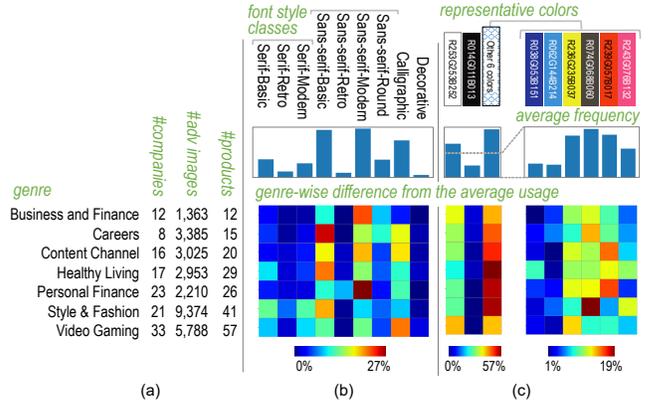}
  \caption{Summary and visualization of font-business relationships in advertisements: a) seven genres and their associated number of images, b) style usage, and c) color usage.}
  \label{fig:hm-adv}
\end{figure}

We classify fonts into one of the nine categories according to a typographer guide~\cite{book2}. We discard unclear fonts that are not listed in the guide or embedded as a bitmap in the file. For colors, we also prepare eight representative colors by the procedure of \ref{sec:fontcolor-book}. The font styles and the representative colors are shown in Fig.\ref{fig:hm-adv}. The frequency of font styles are averaged per product and then per genre.

\subsection{Results}

\subsubsection{Font Styles}
From Fig.~\ref{fig:hm-adv}~(b), we find:
\begin{itemize}
\item Sans-serif is the majority in all genres. Especially, almost all texts are printed in Sans-serif in financial genres: ``Business and Finance'' and ``Personal Finance.''.
\item Serif is preferred in ``Style \& Fashion'', ``Video Gaming'' and ``Healthy Living.'' Especially, ``Style \& Fashion'' is the genre with the least sans-serif frequency.
\item Calligraphy (brush-style) is frequently used in ``Gaming,'' where we find many games on historical matters.
\end{itemize}

\subsubsection{Font Colors}
Fig.~\ref{fig:hm-adv}~(c) reveals the following findings:
\begin{itemize}
    \item Warmer and/or bright colors are majority in all genres.  Especially, red appears in ``Personal Finance'' and ``Content Channel'' and brown in ``Style \& Fashion'' and ``Careers.''
    \item Blue colors are rarely used except that dark blue appears very specifically in ``Healthy Living''.
    \item Interestingly, pink does not correlate with red despite their similarity in hue channel. Red is used in ``Personal Finance'' and ``Content Channel'' as a means of emphasis, whereas pink is used to create an atmosphere in ``Style \& Fashion'' that often targets at women.
\end{itemize}

\subsection{Discussion}
The results clearly suggest that there exist a correlation between fonts and business genres. Compared to book covers, however, online ads have some differences. A good example is that the ratio of Serif in advertisement images are lower than that in book titles. One possible reason is that the resolution of online ads is often low and thus the thinner stroke structure of Serif fonts may damage the visibility. Our analysis not only visualizes the genre and style correlation, but also effectively reveals the differences in media formats.

\section{Conclusion}
In this paper, we quantitatively analyze font usages using large collections of book titles and online advertisements. Our approach successfully identifies interesting trends in terms of how typographic design represents the impression and the atmosphere of the genres. Our cluster analysis using multi-dimensional scaling visualizes that books under close genres share similar font usages; for example, genres for practical books use more sans-serif fonts for their title, while genres for humanity also use serif fonts. Similar trends are found in online advertisements. The discovered trends suggest that typographers carefully choose the style and color of fonts for individual targets such that fonts appropriately represent the content.

Visual font analytics is a novel research direction and has many interesting open problems. In this paper, we only observe the ``average'' usage of each genre; this means that we did not observe the deviation from the average. The deviation is important because visual designs often try to appeal some unexpectedness to impress their audiences. From a marketing science perspective, it is interesting to investigate how fonts and sales correlates, or how fonts affect consumer behavior~\cite{buechel2018buying} in the future. Trends in multiple font combination, text size and position, and visual effects are also open problems for future data-driven typographic study.

\section*{Acknowledgment}
This work was supported by JSPS KAKENHI Grant
Number JP17H06100.


\end{document}